\pdfoutput=1

\documentclass[11pt]{article}

\usepackage[preprint]{acl}

\usepackage{times}
\usepackage{latexsym}

\usepackage[T1]{fontenc}

\usepackage[utf8]{inputenc}

\usepackage{microtype}

\usepackage{inconsolata}

\usepackage{times}
\usepackage{latexsym}
\usepackage{algorithm}
\usepackage{algpseudocode}
\usepackage{amsmath}
\usepackage{amssymb}
\usepackage{amsmath}
\usepackage{arydshln}
\usepackage{ascii}
\usepackage{bm}
\usepackage{booktabs}
\usepackage{colortbl}
\usepackage{color}
\usepackage{CJK}
\usepackage{dsfont}
\usepackage{enumitem}
\usepackage{forest}
\usepackage{graphics}
\usepackage{graphicx}
\usepackage{latexsym}
\usepackage{multirow}
\usepackage{mwe}
\usepackage{makecell}
\usepackage{microtype}
\usepackage{subfig}
\usepackage{subcaption}
\usepackage{times}
\usepackage{tikz}
\usepackage{tabularx}
\usepackage{url}
\usepackage{wrapfig}
\usepackage{tcolorbox}
\usepackage{xcolor}
\usepackage{hyperref}
\usepackage{pifont}

\definecolor{comments}{RGB}{0, 150, 0}

\definecolor{highlight}{RGB}{200, 0, 0}

\definecolor{sred}{RGB}{203, 64, 46}
\definecolor{sblue}{RGB}{44, 73, 135}
\definecolor{sgreen}{RGB}{37, 100, 28}

\definecolor{lightblue}{HTML}{ebf3f8}

\definecolor{mediumblue}{HTML}{d7e8f2}

\definecolor{deepblue}{HTML}{c8dfed}

\title{\includegraphics[width=1.4em]{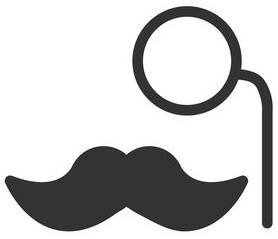}\ Monocle: Hybrid Local-Global In-Context Evaluation for Long-Text Generation with Uncertainty-Based Active Learning}

\author{
\textbf{Xiaorong Wang}\textnormal{\textsuperscript{2}}\thanks{Equal contribution.}\quad
\textbf{Ting Yang}\textnormal{\textsuperscript{3}\footnotemark[1]}\quad
\textbf{Zhu Zhang}\textnormal{\textsuperscript{1}\footnotemark[1]}\quad
\textbf{Shuo Wang}\textnormal{\textsuperscript{1}\thanks{Corresponding authors.}}\\
\textbf{Zihan Zhou}\textnormal{\textsuperscript{4}}\quad
\textbf{Liner Yang}\textnormal{\textsuperscript{5}}\quad
\textbf{Zhiyuan Liu}\textnormal{\textsuperscript{1}}\quad
\textbf{Maosong Sun}\textnormal{\textsuperscript{1}}\footnotemark[2]\\
\\
\textsuperscript{1}Tsinghua University\quad
\textsuperscript{2}Beijing Jiaotong University\\
\textsuperscript{3}Beijing University of Posts and Telecommunications\\
\textsuperscript{4}Xiamen University\quad
\textsuperscript{5}Beijing Language and Culture University\\
}

\begin{document}
\maketitle
\begin{abstract}
Assessing the quality of long-form, model-generated text is challenging, even with advanced LLM-as-a-Judge methods, due to performance degradation as input length increases. To address this issue, we propose a divide-and-conquer approach, which breaks down the comprehensive evaluation task into a series of localized scoring tasks, followed by a final global assessment. This strategy allows for more granular and manageable evaluations, ensuring that each segment of the text is assessed in isolation for both coherence and quality, while also accounting for the overall structure and consistency of the entire piece. Moreover, we introduce a hybrid in-context learning approach that leverages human annotations to enhance the performance of both local and global evaluations. By incorporating human-generated feedback directly into the evaluation process, this method allows the model to better align with human judgment. Finally, we develop an uncertainty-based active learning algorithm that efficiently selects data samples for human annotation, thereby reducing annotation costs in practical scenarios. Experimental results show that the proposed evaluation framework outperforms several representative baselines, highlighting the effectiveness of our approach.
\end{abstract}

\section{Introduction}

Processing longer sequences is a key trend in the development of large language models (LLMs)~\citep{llama-long,data-engineering,ruler,novelqa}. Long-sequence processing can generally be categorized into two main types: long-sequence understanding and long-sequence generation.
Recent advancements in evaluating long-sequence understanding capabilities have made significant progress~\citep{niah,longbench,infinitebench,BABILong,Loong,helmet}, while the evaluation of long-sequence generation tasks remains relatively underexplored.

Existing approaches to long-sequence generation evaluation can generally be categorized into two types: {\em deterministic} and {\em non-deterministic} methods. Deterministic evaluation methods typically break down the overall evaluation into a series of discrete questions~\citep{proxyqa,longgenbench,longproc}, where each question has a deterministic answer (e.g., True or False).
While deterministic evaluation methods can efficiently provide automated scores for long-form model responses, they are often limited to assessing local features and may fail to capture the overall quality of the response. Furthermore, these methods may struggle to generalize to meet the diverse evaluation needs encountered in real-world scenarios.

In contrast to deterministic methods, recent non-deterministic evaluation approaches leverage modern LLMs to directly assess the quality of long-form responses, falling under the category of LLM-as-a-Judge~\citep{surveyllmasajudge}. For example, \citet{longwriter} design an LLM-based evaluator that applies a set of pre-defined principles to generate an overall score for each response. Additionally, HelloEval~\citep{hellobench} introduces a switchable checklist that can be integrated into the LLM-based evaluator, allowing for easy customization across different long-form generation tasks. HelloEval also incorporates human annotations to enhance the reliability of the scores.
However, the reported correlation coefficient between HelloEval and human annotations remains relatively low (0.3193, as noted in \citet{hellobench}), highlighting that the evaluation of long-form generation is still an ongoing challenge.

In this work, we introduce {\em Monocle}, a novel hybrid local-global evaluation framework for long-sequence generation.
Given that the information processing capacity of LLMs can degrade with longer inputs~\citep{infinitebench,longproc}, we adopt a divide-and-conquer approach, splitting the evaluation into two stages: {\em local evaluation} and {\em global evaluation}. In the local evaluation stage, LLMs act as local judges, assessing the quality of smaller chunks derived from the entire response. These local evaluations are then aggregated to form the final score during the global evaluation stage. Additionally, we propose a hybrid in-context learning mechanism to incorporate human annotations. For local demonstrations, we leverage LLMs to generate explanations for the human-assigned local scores, providing deeper insights for the local judges. For global demonstrations, we introduce a concise yet effective structure that includes local scores, local explanations, global score, and global explanations. The global explanations capture key principles for score aggregation or patterns of local-global quality fusion.

Moreover, we propose an uncertainty-based active learning method to enhance the efficiency of human annotation, as relying on human experts to assess the quality of long-form responses is both labor-intensive and time-consuming~\citep{hellobench}. To estimate uncertainty, we follow the approach proposed by \citet{pmlr-v48-gal16}, where we repeatedly sample model scores and calculate the standard deviation as an approximation of the uncertainty.
We combine both local and global uncertainties to guide the selection of data samples for human annotation. By prioritizing samples with higher uncertainty, we focus human annotation efforts on the responses where the model is least confident, which are likely to provide the most informative feedback.

To evaluate the performance of various long-form generation methods, we select paper writing as a representative task, collecting a diverse set of model responses alongside human-assigned quality scores.
Writing a paper from core materials, like experimental results, requires advanced skills from LLMs, including coherent organization, in-depth reasoning, and logical synthesis of information. Unlike creative tasks, paper writing is more objective, making it an ideal benchmark for evaluating the alignment between automatic methods and human judgments.
The resulting dataset, named {\em ReliGen}, serves as a benchmark for the meta-evaluation of long-form generation assessment methods. Experimental results on this dataset demonstrate the effectiveness of the proposed evaluation framework, which outperforms both traditional surface- and semantic-level baselines, as well as modern LLM-as-a-Judge methods.

Our main contributions include:
\begin{itemize}
\item We design a hybrid local-global in-context evaluation method for long-form generation that effectively incorporates human annotations to improve LLM assessments.
\item We propose an uncertainty-based active learning approach to efficiently select data for annotation, enhancing evaluation reliability without increasing the annotation burden.
\item We construct a meta-benchmark to compare the performance of various evaluation methods, with experimental results demonstrating the superiority of our approach.
\end{itemize}

\section{Related Works}

\paragraph{Traditional Evaluation Methods} Conventional metrics for text generation tasks primarily rely on {\em reference-based} evaluation, which requires human-annotated ground-truth texts for quality assessment. At the surface level, BLEU~\citep{bleu} measures the lexical overlap between model responses and references. ROUGE~\citep{rouge} further estimates both precision and recall for model responses. METEOR~\citep{meteor} improves upon exact matches by incorporating synonyms. At the semantic level, some studies calculate the similarity between responses and references in the embedding space, such as BERTScore~\citep{bertscore} and BLEURT~\citep{bleurt}. Unlike these traditional reference-based metrics, which incur high costs due to the need for human-written references, recent studies propose leveraging modern LLMs to directly assess the quality of model responses without requiring ground-truth references.

\paragraph{Deterministic Evaluation Methods} Deterministic evaluation methods typically convert the overall assessment into a series of discrete questions, each with a deterministic answer, to determine whether the generated outputs meet predefined requirements.
LongGenBench~\citep{longgenbench} evaluates long-form generation under complex constraints, assessing how well generated texts adhere to specific details. Similarly, LongProc~\citep{longproc} introduces procedural tasks, such as extracting structured information from HTML to TSV, using rule-based scripts to check the correctness. ProxyQA~\citep{proxyqa} pairs each writing instruction with proxy questions and pre-annotated reference answers. The model-generated response serves as context for answering these questions.
While deterministic evaluation methods automate scoring efficiently, they focus on localized features and often overlook broader aspects like coherence and creativity. Their lack of flexibility can limit their effectiveness in comprehensively assessing model performance.



\paragraph{Non-Deterministic Evaluation Methods} Recent non-deterministic evaluation methods leverage LLMs to directly assess long-form responses. \citet{longwriter} employs an LLM-based evaluator with predefined principles to generate overall scores, while HelloEval~\citep{hellobench} uses 4-6 checklists per task and fits checklist weights via linear regression, incorporating human annotations of overall scores. The effectiveness of HelloEval has been further validated in recent studies~\citep{lcfo}. Recognizing the limitations of LLMs in processing long inputs, we propose a novel divide-and-conquer strategy that incorporates both local and global evaluation. Additionally, we propose an uncertainty-based active learning approach to make better use of human annotations, improving alignment between LLM-based judgments and human evaluations.




\section{Monocle}
\subsection{Problem definition}
\paragraph{Evaluation of Long-Form Generation}
In long-form generation tasks, the model is tasked with generating a long-form response $\mathbf{Y}$ based on the user-provided input $\mathbf{X}$. In this work, we focus on evaluating the quality of the model-generated long sequence $\mathbf{Y}$. Formally, the evaluation process can be represented as
\begin{equation}
    \mathbf{s}=f(\mathbf{X}, \mathbf{Y}),
\end{equation}
where $\mathbf{s}$ represents the quality score of $\mathbf{Y}$, which can be a multi-dimensional vector, with each dimension corresponding to a specific evaluation aspect. The function $f$ denotes the quality estimation method. Recently, there has been growing interest in leveraging LLMs to develop evaluation frameworks~\citep{surveyllmasajudge}.

\begin{figure*}[ht]
  \centering
  \includegraphics[width=0.95\linewidth]{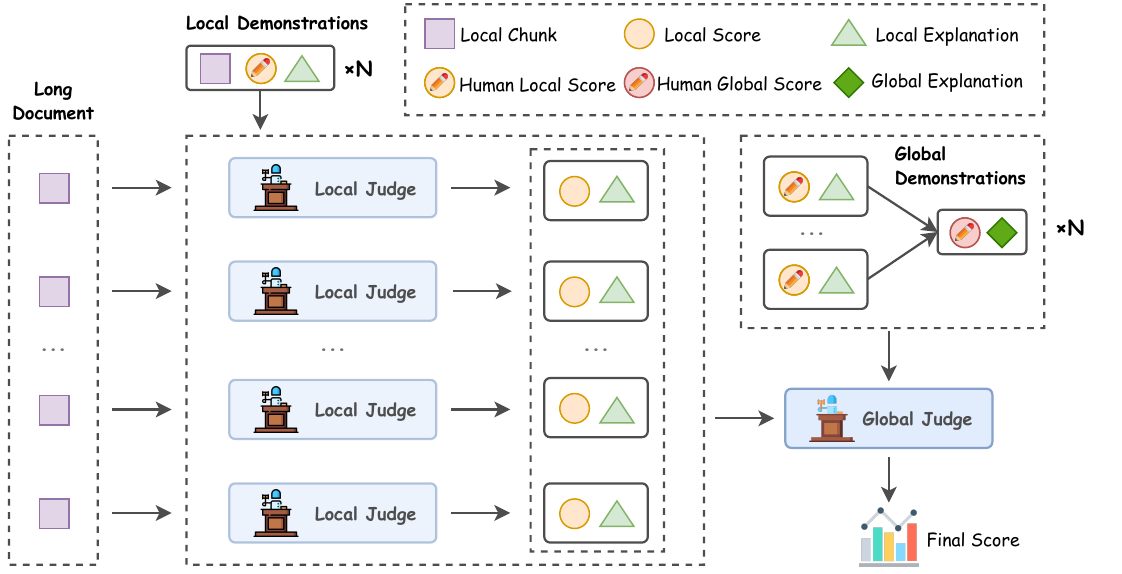}
  \caption{Illustration of the proposed hybrid local-global in-context evaluation framework. Using a divide-and-conquer approach, the evaluation is split into two stages. In the local evaluation stage, LLM-based local judges use demonstrations with human-assigned scores and model-generated explanations to assess the quality of each chunk. In the global evaluation stage, the global judge combines the local scores into a final assessment using score aggregation examples and corresponding global explanations.}
  \label{fig:monocle}
\end{figure*}

\paragraph{In-Context Evaluation} Due to the inherent randomness in the generation process of modern LLMs~\citep{llama,llama2,llama3}, the judgments made by LLMs can vary across different data samples. Additionally, evaluating long-form text is a challenging task that demands advanced capabilities from LLMs, such as long-range information integration and in-depth analysis of logical flow. A promising solution is in-context evaluation, which utilizes human-annotated scores as demonstrations to enhance the alignment between LLM-generated scores and human assessments.
Formally, let $d = \langle \mathbf{X}^{\prime}, \mathbf{Y}^{\prime}, \mathbf{s}^{\prime} \rangle$ be the human-scored demonstration. We define $\mathcal{D} = \{ d^{(n)} \}_{n = 1}^{N}$ as the set of all human-annotated demonstrations, where $N$ is the number of demonstrations (i.e., shots). The formal representation can be given by
\begin{equation}
\mathbf{s} = f^{\mathrm{ICL}}\left(\mathbf{X}, \mathbf{Y}, \mathcal{D} \right).
\end{equation}

\subsection{Hybrid In-context Evaluation}
Previous studies have shown that the information processing capabilities of most LLMs tend to degrade with longer inputs~\citep{infinitebench,longproc}, even though they can perform well on certain simplified long-sequence tasks (e.g., needle-in-the-haystack~\citep{niah}). To address this, we propose adopting the idea of divide-and-conquer~\citep{llmxmapreduce} to construct a hybrid local-global evaluation framework. Additionally, we develop a specific in-context learning approach that is well-suited for this hybrid local-global framework.


\paragraph{Chunking} To facilitate the divide-and-conquer evaluation process, we split the long-form output $\mathbf{Y}$ into several chunks based on task-specific properties. For example, in the paper writing task, each chunk corresponds to one or more sections. The input $\mathbf{X}$ is also divided to align with the output. For instance, the materials provided for the $j$-th chunk, $\mathbf{y}_j$, in $\mathbf{Y}$ are used to derive the corresponding input, $\mathbf{x}_j$. The chunking process can be represented as
\begin{equation}
\begin{split}
    \mathbf{X} &= \{x_{1}, \cdots, x_{J}\}, \\
    \mathbf{Y} &= \{y_{1}, \cdots, y_{J}\},
\end{split}
\end{equation}
where $J$ is the number of chunks for $\mathbf{X}$ and $\mathbf{Y}$.

\paragraph{Local In-context Evaluation} In this stage, each local chunk undergoes a dedicated assessment process, supported by a set of local demonstrations that provide additional context. Formally, each local demonstration is represented as a structured tuple:
\begin{equation}
d^{\mathrm{local} (n)} = \langle \mathbf{y}^{\prime (n)}, \mathbf{l}^{\prime (n)}, \mathbf{q}^{\prime (n)} \rangle,
\end{equation}
where $n$ indexes the $n$-th demonstration within the set $\mathcal{D}^{\mathrm{local}}=\{d^{\mathrm{local} (n)}\}_{n=1}^{N}$. $\mathbf{y}^{\prime (n)}$ denotes the extracted chunk from the exemplary model-generated response, while $\mathbf{l}^{\prime (n)}$ is the corresponding local score assigned by human evaluators. $\mathbf{q}^{\prime (n)}$ represents the associated local explanation that clarifies the rationale behind the score. Note that $\mathbf{q}^{\prime (n)}$ is generated by LLMs, which significantly reduces the cost of human annotation.
Once the local demonstrations are established as in-context exemplars, the local quality assessment for a given chunk $\mathbf{y}_j$ can be formulated as follows:
\begin{equation}
      \mathbf{l}_j, \mathbf{q}_j = f^{\mathrm{local}}(\mathbf{x}_j, \mathbf{y}_j, \mathcal{D}^{\mathrm{local}}),
\end{equation}
where $f^{\mathrm{local}}$ represents the local judge driven by LLMs. As shown in Figure~\ref{fig:monocle}, the local judge generates both the local score $\mathbf{l}_j$ and the corresponding local explanation $\mathbf{q}_j$, which will be aggregated in the subsequent global evaluation stage.

\paragraph{Global In-context Evaluation} This stage is designed to synthesize the local assessments into a comprehensive global evaluation, taking into account both the pre-estimated local scores and global patterns, such as structure and logical flow.
The main challenge in this stage is designing the structure of the global demonstrations. Using full-length responses directly would lead to excessively long inputs for the global judge. For example, if the average length of a full-length demonstration is $L$, the total input length would exceed $N \times L$, potentially overwhelming the evaluation capabilities of the global judge. We thus propose a concise and effective structure for the global demonstrations:
\begin{equation}
\begin{split}
    d^{\mathrm{global} (n)} = \langle \{(\mathbf{l}^{\prime (n)}_i, \mathbf{q}^{\prime (n)}_i) \}_{i=1}^{I^{(n)}}, 
    \mathbf{s}^{\prime (n)}, \mathbf{a}^{\prime (n)} \rangle, \\
\end{split}
\label{eq:global-demo}
\end{equation}
where $d^{\mathrm{global} (n)}$ represents the $n$-th global demonstration in the set $\mathcal{D}^{\mathrm{global}}=\{d^{\mathrm{global} (n)}\}_{n=1}^{N}$. $I^{(n)}$ denotes the number of chunks in the response for the $n$-th shot. Each global demonstration include:
\begin{itemize}
    \item $\{(\mathbf{l}^{\prime (n)}_i, \mathbf{q}^{\prime (n)}_i) \}_{i=1}^{I^{(n)}}$ contains the local evaluation results, where each item consists of a local human-assigned quality score $\mathbf{l}^{\prime (n)}_i$ and its corresponding local explanation $\mathbf{q}^{\prime (n)}_i$, both derived from the $i$-th chunk of the $n$-th shot. 
    \item $\mathbf{s}^{\prime (n)}$ signifies the human-assigned global quality score for the entire model response in the $n$-th shot. This score reflects an overall assessment, taking into account the synthesis of all constituent chunks and global patterns.
    \item $\mathbf{a}^{\prime (n)}$ represents the model-generated global explanation, which is based on the local responses $\{ \mathbf{y}_i \}_{i=1}^{I^{(n)}}$, local human-assigned scores $\{ \mathbf{l}^{\prime (n)}_i \}_{i=1}^{I^{(n)}}$, and the global human-annotated score $\mathbf{s}^{\prime (n)}$. This explanation provides a comprehensive rationale for the global score $s^{\prime}_n$, highlighting overarching strengths and weaknesses across all chunks. It explains how the local qualities contribute to the overall global quality and how local scores are fused with global features. The global explanation is intended to assist the global judge in more accurately estimating the final score for the current test example $\mathbf{Y}$.
\end{itemize}

As shown in Figure~\ref{fig:monocle}, the global judge uses global demonstrations, as illustrated in Eq.~(\ref{eq:global-demo}), along with the local responses and their evaluated results, to determine the final score:
\begin{equation}
    \mathbf{s} = f^{\mathrm{global}}(\{\mathbf{y}_j, \mathbf{l}_j, \mathbf{q}_j\}_{j=1}^{J}, \mathcal{D}^{\mathrm{global}}).
\end{equation}

\subsection{Uncertainty-based Active Learning}
\label{sec:uncertainty}
For long-form generation tasks, manually annotating large volumes of data is an inherently time-consuming and labor-intensive process that requires significant human effort and incurs high costs. To address this challenge, we propose an uncertainty-based active learning strategy, where the model actively selects and prioritizes data points that are most ambiguous or difficult to evaluate (i.e., data samples with high uncertainty). This focused approach helps optimize the use of human resources, reducing the overall annotation burden while improving model performance.

\begin{figure}[h]
\centering
\includegraphics[width=0.95\linewidth]{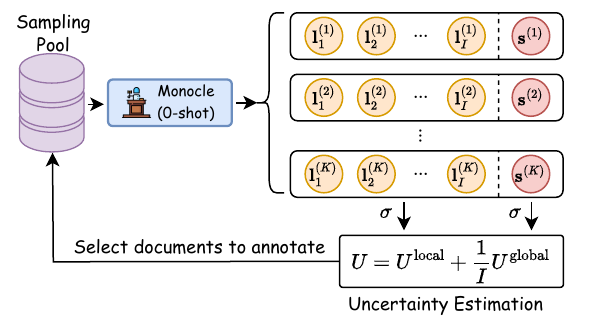}
\caption{Illustration of the proposed uncertainty-based activate learning method.}
\label{fig:uncertainty}
\end{figure}

\paragraph{Uncertainty Estimation} For data samples where determining a score is challenging, the model may exhibit uncertainty in its prediction. As illustrated in Figure~\ref{fig:uncertainty}, we quantify this uncertainty by repeatedly applying the proposed Monocle framework for evaluation. We then use the standard deviation of the local and global scores to represent the local and global uncertainties, respectively. Quantifying model uncertainty through repeated sampling is a widely used approach in previous studies~\citep{pmlr-v48-gal16,xiao-etal-2019-uncertainty,wang-etal-2019-uncertainty}. Intuitively, when the model lacks confidence in its prediction, the predicted scores tend to vary over a larger range, resulting in higher uncertainty. Note that during this procedure, Monocle operates in a zero-shot manner, eliminating the influence of in-context demonstrations on the inherent uncertainty of the model.


\paragraph{Active Data Selection} Given a sampling pool $\mathcal{P}$ containing a large number of unlabeled data points, our goal is to select a subset of $N$ data points for human annotation.
We perform data selection at the chunk level. To account for both local and global uncertainties, we combine these two types of uncertainty for each chunk:
\begin{equation}
    U = U^{\mathrm{local}} + \frac{1}{I}U^{\mathrm{global}},
\end{equation}
where $U^{\mathrm{local}}$ represents the uncertainty derived from the sampled local scores $\{ \mathbf{l}_1^{(k)}, \cdots, \mathbf{l}_{I}^{(k)} \}_{k=1}^{K}$, while $U^{\mathrm{global}}$ denotes the global uncertainty calculated based on the sampled global scores $\{ \mathbf{s}^{(k)} \}_{k=1}^{K}$. $K$ is the number of repetitions. Subsequently, we select the top-$N$ data samples with the highest uncertainty scores for human annotation.

\section{ReliGen}
\paragraph{Principles} To evaluate the reliability of methods for assessing long-form generation, the benchmark should: (1) include both writing instructions and reference materials, allowing for comparison between reference-based and reference-free approaches; (2) incorporate human-assigned scores to assess the correlation between machine-predicted and human-assigned evaluations; and (3) feature tasks that mirror real-world applications of long-form generation, ensuring practical relevance.\footnote{To the best of our knowledge, only \citet{hellobench} have constructed both long-form responses and human-assigned scores, but their human-evaluated results have not been open-sourced and are not publicly available.}

We choose paper writing as the task for our meta-evaluation benchmark because it demands advanced capabilities from LLMs, including organizing material coherently, in-depth reasoning, and synthesizing information logically. Unlike more subjective tasks like novel writing, paper quality is relatively objective, making it ideal for assessing the alignment between automatic evaluation methods and human judgments. By focusing on this task, we can better understand how well different evaluation strategies capture key elements of coherence, structure, and critical analysis that are crucial to long-form generation.


\begin{figure}[ht]
  \centering
  \includegraphics[width=0.8\linewidth]{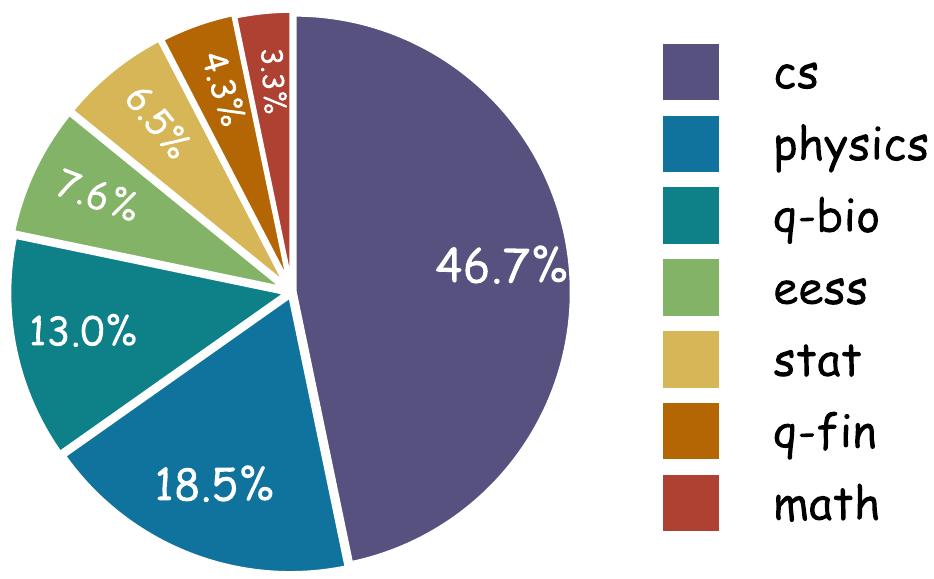}
  \caption{Distribution of papers in ReliGen.}
  \label{fig:data_distribution}
\end{figure}

\subsection{Data Collection}
\label{sec:data-construct}

To ensure the quality and diversity of ReliGen, we first collect publicly available papers from arXiv\footnote{\url{https://arxiv.org}}. These papers are then used to create writing instructions through back-translation and human verification. Using the verified instructions, we employ various LLMs and decoding strategies to generate a diverse set of model responses.

\paragraph{Reference Papers} Figure~\ref{fig:data_distribution} presents the topic distribution of the reference papers collected in ReliGen. By ensuring topic diversity, the benchmark is designed to better capture the evaluation capabilities of various approaches, making the results more robust and generalizable to real-world applications. Each paper is in the LaTex format.

\paragraph{Writing Instructions} To generate high-quality writing instructions that are closely aligned with the references, we follow a two-step process: back-translation and human verification.
Please refer to Section~\ref{appendix-sec:instruction} in Appendix for more details.

\paragraph{Model Responses} In accordance with human-verified writing instructions, we employ diverse LLMs and decoding strategies to generate the corresponding responses. Please refer to Section~\ref{appendix-sec:response} in Appendix for more details.

After eliminating invalid outputs, the final dataset consists of 92 entries, each containing a writing instruction, a model response, and a reference paper. For the length distribution of both the reference papers and model responses, please refer to Figure~\ref{appendix-fig:length_distribution} in Appendix.

\subsection{Manual Scoring}
\label{sec:human-annotate}
To assess the reliability of various evaluation methods for long-form generation, we employ a comprehensive and systematic manual scoring process to establish ground-truth scores for the model responses, as detailed in Section~\ref{sec:data-construct}. This approach allows us to rigorously compare the performance of the evaluation methods. Our primary objective is to analyze the correlation between the predicted scores, derived from these evaluation frameworks, and the human-assigned ground-truth scores. By examining this correlation, we aim to determine the effectiveness of each evaluation method in accurately reflecting the quality of generated content. This evaluation provides crucial insights into the strengths and limitations of different scoring approaches in the context of long-form generation. Please refer to Section~\ref{appendix-sec:score} in Appendix for more details of the human annotation process.

\section{Experiments}
\subsection{Setup}
\paragraph{Models} Due to the cost-effectiveness of GPT-4o-mini, we use this model by default for ablation studies to investigate the impact of each component in Monocle. Additionally, we evaluate the performance of Monocle using a range of representative LLMs, including GPT-4o, Claude-3.5-Sonnet, and Gemini-1.5-Pro, in the following experiments. During inference, the default decoding temperature for all models is set to 0.1, except during the uncertainty estimation process, where the temperature is increased to 0.8 to enhance the diversity.

\paragraph{Evaluation}
To evaluate the performance of various assessment metrics and validate their alignment with human judgment, we conducted a series of experiments on the carefully annotated ReliGen dataset. To measure the correlation, we used the \texttt{scipy}\footnote{\url{https://scipy.org}} package to compute the Spearman rank correlation coefficient, denoted as $\rho$, and the associated p-value, $p$.


\subsection{Ablation Study}

\begin{table}[ht]
\centering
\begin{tabular}{l c cc}
    \toprule
    \bf Method & \bf ICL & $\bm{\rho}$ ($\uparrow$) & $\bm{p}$ ($\downarrow$) \\
    \cmidrule(lr){1-2} \cmidrule(lr){3-4}
    \multirow{2}{*}{Direct Judge}  & \ding{55} & 0.228 & 2.83$\times$10$^{-2}$ \\
    & \ding{51}  & 0.205 & 4.91$\times$10$^{-2}$ \\
    \cmidrule(lr){1-2} \cmidrule(lr){3-4}
    \multirow{2}{*}{Hybrid Judge}  & \ding{55} & 0.351 & 5.96$\times$10$^{-4}$ \\
    & \ding{51}  & \bf 0.550 & 1.34$\times$10$^{-8}$ \\
    \bottomrule
\end{tabular}
\caption{Effect of hybrid in-context evaluation.}
\label{tab:hybrid_local_global}
\end{table}

\paragraph{Effect of Hybrid In-Context Evaluation} As mentioned earlier, we adopt a divide-and-conquer approach to construct a hybrid local-global evaluation framework, aiming to alleviate the challenge that the comprehensive capabilities of LLMs may decrease with longer inputs~\citep{infinitebench}. Table~\ref{tab:hybrid_local_global} presents the results, which demonstrate that the proposed hybrid evaluation method outperforms direct LLM-as-a-Judge methods~\cite{longwriter,hellobench}, even in a zero-shot setting (0.351 vs. 0.228). Furthermore, when provided with in-context, human-annotated examples selected using the same method, our hybrid approach achieves a higher correlation with human assessments (0.550). These results demonstrate that our approach not only leverages LLMs more effectively for long-form evaluation but also makes better use of human annotations.



\begin{table}[ht]
\centering
\begin{tabular}{l cc}
    \toprule
    \bf Demonstration & $\bm{\rho}$ ($\uparrow$) & $\bm{p}$ ($\downarrow$) \\
    \cmidrule(lr){1-1} \cmidrule(lr){2-3}
    None & 0.351 & 5.96$\times$10$^{-4}$ \\
    \cmidrule(lr){1-1} \cmidrule(lr){2-3}
    Local & 0.479 & 1.28$\times$10$^{-6}$  \\
    Local + Global & \bf 0.550 & 1.34$\times$10$^{-8}$ \\
    \bottomrule
\end{tabular}
\caption{Effect of local and global demonstrations.}
\label{tab:effect_few_shot}
\end{table}
\paragraph{Effect of Local-Global Demonstrations}
Table \ref{tab:effect_few_shot} presents an investigation into the impact of incorporating demonstrations at different stages of the evaluation process. The results show that both local and global demonstrations effectively enhance the alignment between model-predicted scores and human-assigned scores, reaffirming the importance of the hybrid evaluation approach in Monocle.



\begin{table}[ht]
\centering
\begin{tabular}{l cc}
    \toprule
    \bf Selection Method & $\bm{\rho}$ ($\uparrow$) & $\bm{p}$ ($\downarrow$) \\
    \cmidrule(lr){1-1} \cmidrule(lr){2-3}
    None & 0.351 & 5.96$\times$10$^{-4}$ \\
    \cmidrule(lr){1-1} \cmidrule(lr){2-3}
    Random & 0.389 & 1.23$\times$10$^{-4}$ \\
    PPL-based & 0.451 & 6.24$\times$10$^{-6}$  \\
    Uncertainty-based & \bf 0.550 & 1.34$\times$10$^{-8}$ \\
    \bottomrule
\end{tabular}
\caption{Effect of different data selection methods for human-annotated demonstrations.}
\label{tab:example_selection}
\end{table}

\paragraph{Effect of Uncertainty-Based Active Learning}
To thoroughly validate the effectiveness of the proposed uncertainty-based active learning approach, we compare different data selection methods for human annotation.
Each data sample includes a writing instruction, a model response, and a reference. The selected samples are then provided to human annotators, who assign quality scores to the responses, following the same annotation process outlined in Section~\ref{sec:human-annotate}. Note that there is no overlap between the annotations used for in-context demonstrations and those for the test examples.
We investigate the performance of the following three selection methods:
\begin{itemize}
    \item {\em Random Selection}: Randomly select $N$ data samples for annotation.
    \item {\em PPL-based Selection}: Use perplexity (PPL) as a criterion for example selection, with data samples having higher PPLs being more likely to be chosen.
    \item {\em Uncertainty-based Selection}:  Use the proposed uncertainty-based active learning method described in Section~\ref{sec:uncertainty}.
\end{itemize}

The results in Table~\ref{tab:example_selection} show clear differences in the effectiveness of these strategies. The random selection method leads to no significant performance gains. In contrast, both the PPL-based and uncertainty-based strategies can outperform the random baseline, with the uncertainty-based strategy achieving the highest correlation and lowest p-value. This suggests that selecting examples based on the model uncertainty in scoring provides a more effective approach, greatly improving the evaluation performance without increasing human-annotation burden.

\begin{table}[ht]
\centering
\begin{tabular}{l cc}
    \toprule
    \bf Model & $\bm{\rho}$ ($\uparrow$) & $\bm{p}$ ($\downarrow$) \\
    \cmidrule(lr){1-1} \cmidrule(lr){2-3}
    GPT-4o-mini         & 0.550     & 1.34$\times$10$^{-8}$ \\
    GPT-4o              & 0.479     & 1.33$\times$10$^{-6}$  \\
    Claude-3.5-Sonnet   & \bf 0.568 & 3.30$\times$10$^{-9}$ \\
    Gemini-1.5-Pro      & 0.540     & 2.73$\times$10$^{-8}$ \\
    \bottomrule
\end{tabular}
\caption{Compatibility of Monocle with different LLMs.}
\label{tab:differen_model}
\end{table}

\paragraph{Compatibility with Different Models} We validated the effectiveness of our method across different models, as shown in Table \ref{tab:differen_model}. The experiments involved four models: GPT-4o-mini, GPT-4o, Claude-3.5-Sonnet, and Gemini-1.5-Pro. All models showed strong alignment with human ratings using our method. For the subsequent main experiments, we will select Claude-3.5-Sonnet as the backbone model for both our method and the HelloEval~\citep{hellobench} baseline.

\subsection{Main Results}

\begin{table}[t]
\setlength{\tabcolsep}{4.5pt}
\centering
\begin{tabular}{l cc}
    \toprule
    \bf Methods & $\bm{\rho}$ ($\uparrow$) & $\bm{p}$ ($\downarrow$) \\
    \midrule
    \multicolumn{3}{c}{\cellcolor{lightblue}{\em Surface-Level Evaluation}} \\
    BLEU~{\tiny\citep{bleu}} & 0.148 & 1.56$\times$10$^{-1}$ \\
    ROUGE-L~{\tiny\citep{rouge}} & 0.201 & 5.35$\times$10$^{-2}$ \\
    METEOR~{\tiny\citep{meteor}} & 0.206 & 4.79$\times$10$^{-2}$ \\
    ChrF++~{\tiny\citep{chrf++}} & 0.231 & 2.65$\times$10$^{-2}$ \\
    \midrule
    \multicolumn{3}{c}{\cellcolor{mediumblue} \em Semantic-Level Evaluation} \\
    BERTScore~{\tiny\citep{bertscore}} & 0.151 & 1.50$\times$10$^{-1}$ \\
    BLEURT~{\tiny\citep{bleurt}} & 0.013 & 9.01$\times$10$^{-1}$ \\
    \midrule
    \multicolumn{3}{c}{\cellcolor{deepblue} \em LLM-Based Evaluation} \\
    HelloEval~{\tiny\citep{hellobench}} & 0.275 & 7.91$\times$10$^{-3}$ \\
    Monocle~{\tiny (\textbf{Ours})} & \bf 0.568 & 3.30$\times$10$^{-9}$ \\
    \bottomrule
\end{tabular}
\caption{Performance of various evaluation methods on the meta-evaluation benchmark ReliGen for the long-form generation task.}
\label{tab:benchmakink_metrics}
\end{table}

\begin{figure*}[t]
    \centering
    \subfloat[Correlation between Monocle and human evaluation.]{
    \includegraphics[height=0.235\textwidth]{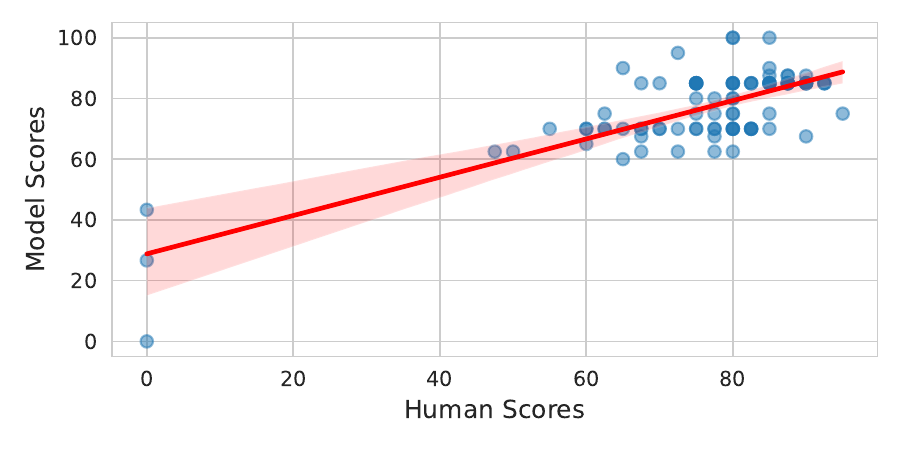}}
    \hspace{0.03\textwidth}
    \subfloat[Correlation between HelloEval and human evaluation.]{
    \includegraphics[height=0.235\textwidth]{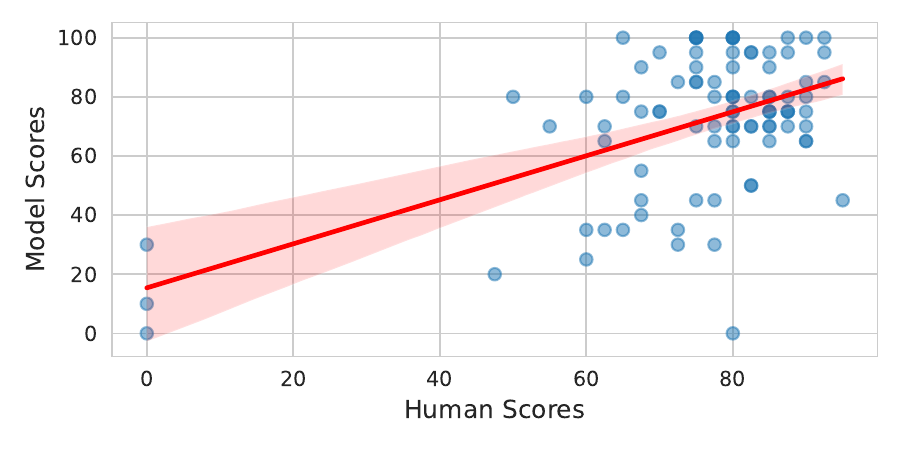}}

    \subfloat[Score distribution of Monocle]{
    \includegraphics[height=0.235\textwidth]{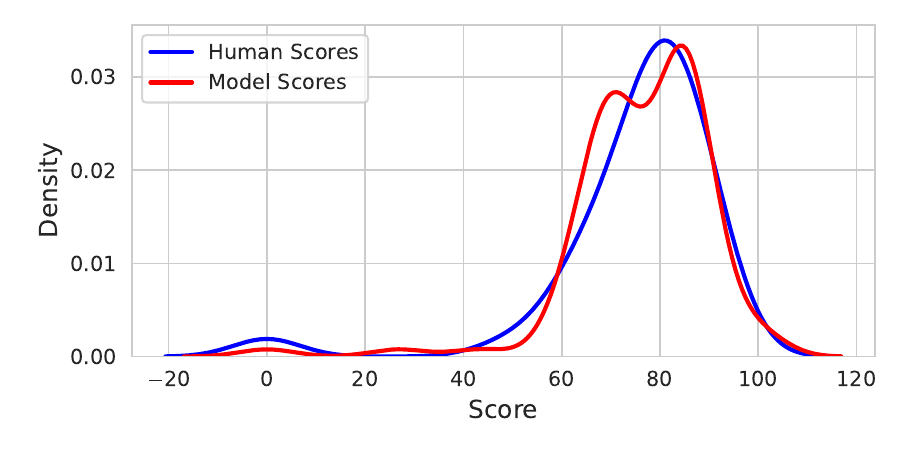}}
    \hspace{0.03\textwidth}
    \subfloat[Score distribution of HelloEval]{
    \includegraphics[height=0.235\textwidth]{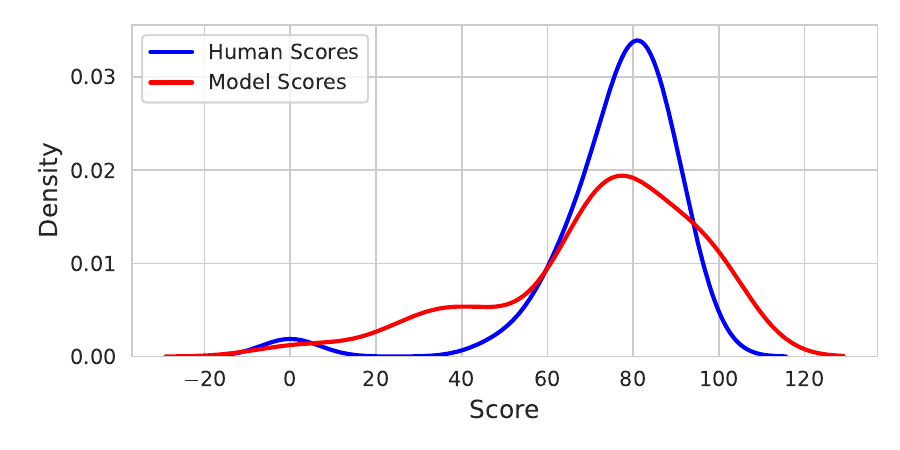}}
\caption{Visualization of the scores assigned by both Monocle and the HelloEval baseline, where both methods incorporate a small amount of human annotation to enhance the reliability of the evaluation results.}
\label{fig:score_vis}
\end{figure*}





We compare our method with several representative and widely-used baselines, of which the implementation details are included in Section~\ref{appendix-sec:baseline} in Appendix.
As shown in Table~\ref{tab:benchmakink_metrics}, both surface- and semantic-level evaluation metrics generally exhibit low correlations with human judgments, indicating their potential limitations in assessing long-form text. While HelloEval~\citep{hellobench}, an LLM-as-a-Judge approach, demonstrates a notable improvement in correlation compared to previous metrics, its direct processing of lengthy model-generated responses limits its ability to capture finer details. In contrast, our Monocle method, which employs a hybrid local-global strategy, achieves a correlation of 0.568, reflecting superior performance. It indicates a strong positive correlation when Spearman’s $\rho$ is greater than 0.5~\citep{autosurvey}. Figure~\ref{fig:score_vis} presents a visualization of the scores. It's worth noting that HelloEval also leverages human-annotated assessments as examples for linear regression, whereas we integrate the same human-annotated examples through the proposed hybrid local-global in-context evaluation. Moreover, since our method is reference-free, it can be easily applied in more practical scenarios, without the need for human experts to create long-form references. We also investigate the effect of the number of demonstrations in Section~\ref{appendix-sec:shot} in Appendix.


\section{Conclusion}

We introduce Monocle, a hybrid local-global in-context evaluation framework for long-form generation. Through experiments, we demonstrate the effectiveness of each component in Monocle. Furthermore, our results show that Monocle outperforms a wide range of representative baselines.

\section*{Limitations}

In this work, we evaluate the proposed method in a training-free setting. It remains unclear whether the performance could be further enhanced by training the backbone model to learn domain-specific evaluation capabilities. Additionally, due to the prohibitively high cost of collecting human assessments for long-form generation across various tasks, we focus only on the correlation of the methods with human scores in the paper writing scenario. Future work will extend this approach to more practical scenarios and domains, providing a more comprehensive evaluation.



\bibliography{Main.bbl}

\appendix
\clearpage

\section{Length Distribution of ReliGen}

\begin{figure}[ht]
    \centering
    \subfloat[Length distribution of reference papers.]{
    \includegraphics[height=0.235\textwidth]{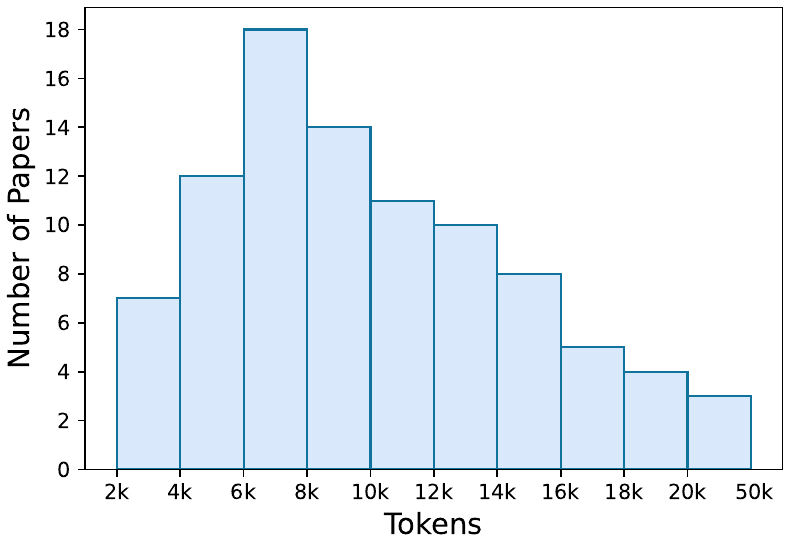}}
    
    \subfloat[Length distribution of model responses.]{
    \includegraphics[height=0.235\textwidth]{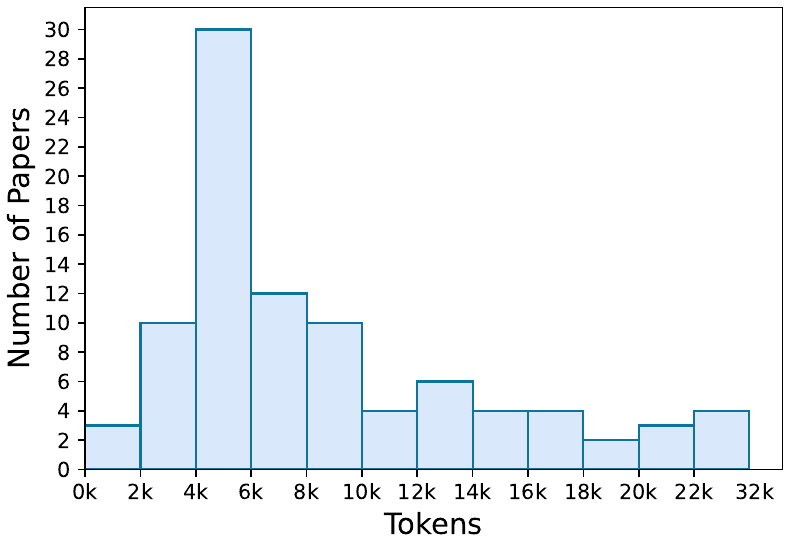}}
\caption{Length distribution of the reference papers and the model responses in ReliGen.}
\label{appendix-fig:length_distribution}
\end{figure}

Figure~\ref{appendix-fig:length_distribution} illustrates the detailed length distribution of both the reference papers and the model responses in ReliGen.

\section{Details of Writing Instructions in ReliGen}
\label{appendix-sec:instruction}
First, we use LLM-based back-translation to create instructions by extracting key equations, experimental results, and core intuitions from the reference material. This ensures that the essential components of the instructions are consistent with the references, eliminating the need for the model to independently generate objective content like experimental results. Next, each instruction undergoes validation by two human reviewers, focusing on faithfulness, clarity, and adherence to LaTeX grammar. Any redundant content or errors are corrected to ensure the instructions are accurate and clear.

\section{Details of Responses in ReliGen}
\label{appendix-sec:response}
Generating lengthy responses exceeding 10,000 tokens presents a significant challenge. To address this, we implemented two distinct generation methods: standard decoding, which uses LLMs to generate responses directly from the instructions, and AgentWrite~\citep{longwriter}, which generates responses in a section-by-section manner. To enhance the quality and diversity of the generated responses, we utilized six different models, including both closed- and open-source options: GPT-4o, GPT-4o-mini, GPT-o1-mini, Claude-3.5-Sonnet, LLaMA-3.1-8B, and Qwen-2.5-7B. In the final dataset, 80\% of the data was generated using AgentWrite, with the remaining 20\% generated via standard decoding.

\section{Details of Human Annotation}
\label{appendix-sec:score}

\paragraph{Annotator Recruitment} We recruited a diverse team of 15 human annotators, consisting of both undergraduate and graduate students, to ensure a broad range of perspectives and expertise. Each annotator was compensated 50 RMB per data point, with the average time spent annotating each instance being approximately one hour. A subset of these annotators had prior experience with NLP annotation tasks, providing a solid foundation in data labeling and enhancing the overall quality of the annotations. To further strengthen the reliability and objectivity of the evaluation process, we intentionally selected annotators from a variety of academic disciplines, such as Computer Science, Linguistics, and English Translation. This multi-disciplinary approach was designed to mitigate potential biases and encourage diverse viewpoints. Additionally, all annotators demonstrated strong English reading comprehension skills, which were essential for accurately assessing the dataset.


\paragraph{Details of Annotation} We provide both the model responses and their corresponding reference papers to the human evaluators. However, evaluators are explicitly instructed that the reference papers are intended as a guide, and model responses are not required to match them exactly.
For each data point, we adopted a double-annotation strategy, assigning two independent annotators to evaluate the response. To preserve objectivity and prevent any potential biases, the results from the two annotators were kept blind from one another. In cases where significant discrepancies arose between the initial annotations, a third expert evaluator was brought in to provide a final, adjudicated assessment, ensuring a more robust and reliable evaluation process.
Furthermore, the annotation task employed a two-tiered scoring system, with annotators required to provide scores at both the section level and the overall paper level. For each data point, the final section scores and the overall paper score were determined by averaging the ratings from all the involved annotators.

\section{Details of Baselines}
\label{appendix-sec:baseline}
We compare our method against several representative baselines. The implementation details are provided below:
\begin{itemize}
    \item {\em BLEU}: Use the \texttt{sacrebleu}\footnote{\url{https://github.com/mjpost/sacrebleu}} package to calculate the BLEU score between model responses and the references.
    \item {\em ROUGE-L}: Use the \texttt{rouge\_score}\footnote{\url{https://pypi.org/project/rouge-score}} package to calculate the ROUGE-L score between model responses and the references.
    \item {\em METEOR}: Use the \texttt{nltk}\footnote{\url{https://www.nltk.org/}} package to calculate the METEOR score between model responses and the references.
    \item {\em ChrF++}: Use the \texttt{sacrebleu}\footnote{\url{https://github.com/mjpost/sacrebleu}} package to calculate the ChrF++ score between model responses and the references.
    \item {\em BERTScore}: Use the \texttt{bert\_score}\footnote{\url{https://github.com/Tiiiger/bert_score}} package to calculate the BERTScore between model responses and references, based on \texttt{ModernBERT}\footnote{\url{https://github.com/AnswerDotAI/ModernBERT}}.
    \item {\em BLEURT}:  Use the \texttt{bleurt}\footnote{\url{https://github.com/google-research/bleurt}} package to calculate the BLEURT score between model responses and references, based on \texttt{BLEURT-20}\footnote{\url{https://storage.googleapis.com/bleurt-oss-21/BLEURT-20.zip}}.
    \item {\em HelloEval}: Slightly adapt the prompt from \citet{hellobench} to suit the paper writing task. Use the code open-sourced by the authors\footnote{\url{https://github.com/Quehry/HelloBench}} to perform linear regression with human-annotated examples. To ensure a fair comparison, the same human assessments are used for both HelloEval and Monocle.
\end{itemize}

\section{Analysis of the Number of Demonstrations}
\label{appendix-sec:shot}

\begin{figure}[h]
\centering
\includegraphics[width=0.95\linewidth]{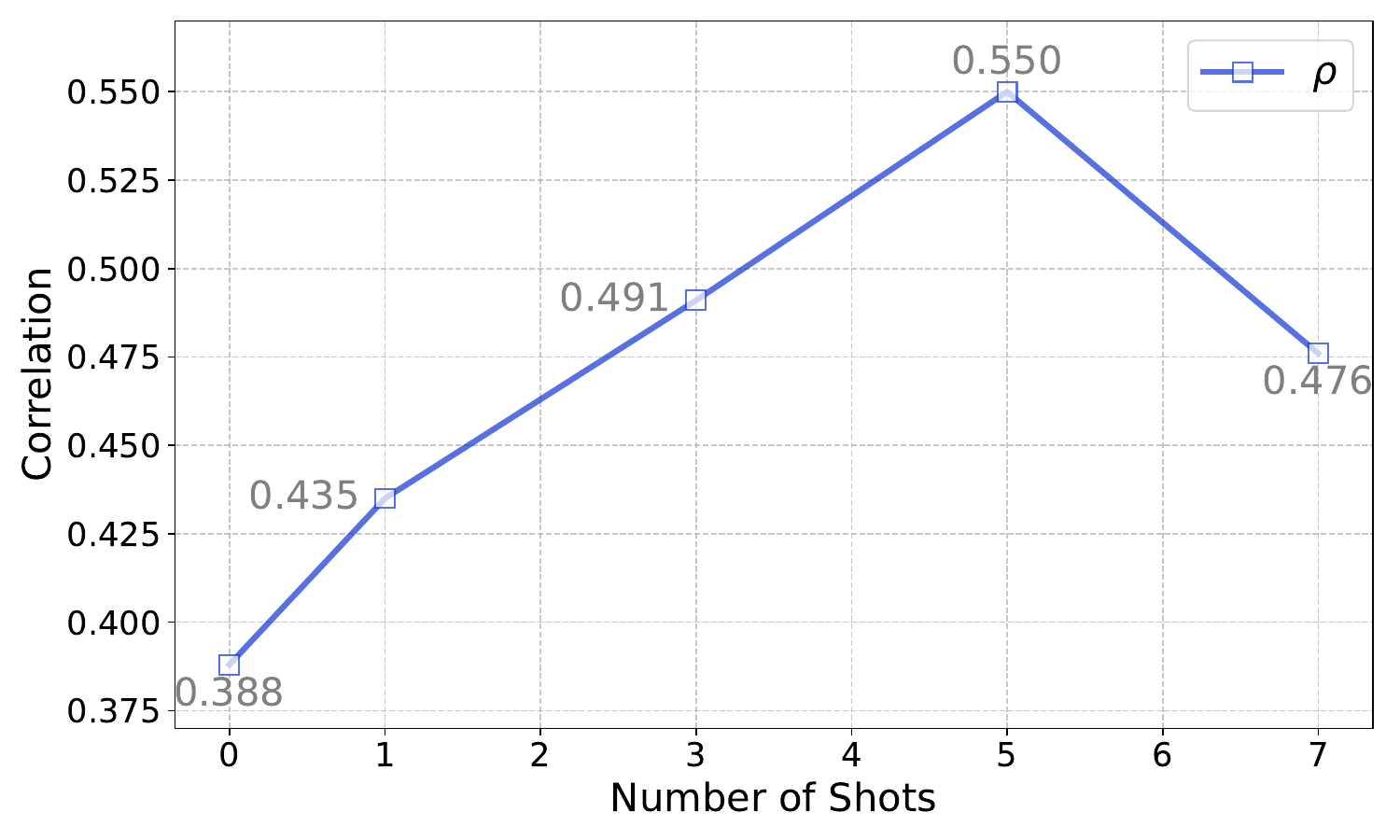}
\caption{Performance when using different numbers of demonstrations.}
\label{fig:shot_num}
\end{figure}

Figure~\ref{fig:shot_num} analyzes the relationship between the number of shots and the Spearman’s rank correlation coefficient $\rho$. The results show that correlation improves with more shots, peaking at an optimal value of five demonstrations. Beyond this point, additional shots offer diminishing returns, emphasizing the importance of selecting an appropriate number of demonstrations to balance informativeness and processing efficiency for optimal evaluation accuracy.

\section{Detailed Prompts}

To better present the overall process of Monocle, we provide the prompts used. Specifically, Figure \ref{fig:generate_local_explanatin} shows the prompt that guides the model to generate local explanation. Figure \ref{fig:prompt_for_agg_exp} demonstrates the prompt that instructs the model to aggregate local scores and generate global explanation. Figure \ref{fig:prompt_for_local_judge} presents the prompt for the local in-context evaluation stage, and Figure \ref{fig:prompt_for_aggregation} presents the prompt for the global in-context evaluation stage.


\begin{figure*}[h]
 \begin{tcolorbox}[colback=blue!2,colframe=blue!50!black]
 \tiny
 \texttt{You are a helpful AI assistant.}\\
 \texttt{Your task is to explain the human-assigned rating for a model-generated paper section.}\\
 \texttt{Below are the Judge Guidelines used by human evaluators to score the model-generated sections.}\\
 \texttt{Please read these guidelines carefully, then see the original input, model generated response, and the human rating. Based on these guidelines, provide a concise yet specific explanation of the human-assigned rating, focusing on relevant dimensions.}\\
 \texttt{--------------------------------------------------------------------------------}\\
 \texttt{**Judge Guidelines:**}\\
 \texttt{When assessing the quality of the model-generated section, please consider the following dimensions:}
 \texttt{[Evaluation Dimensions]}\\
 \texttt{1. **Structure and Format**:}\\
   \texttt{- Does the generated paper’s section follow standard academic structure?}\\
   \texttt{- Are numbering, subsection titles, and hierarchy within the section clear?}\\
   \texttt{- Is LaTeX formatting (e.g., tables, formulas, citations) correct?}\\
\texttt{2. **Content Coverage**:}\\
   \texttt{- Does the section comprehensively cover all the essential content specified by the input materials?}\\
\texttt{3. **Detail and Depth**:}\\
   \texttt{- Does the section provide enough detailed information to explain its purpose and content fully?}\\
   \texttt{- Are all relevant aspects of the topic explored in sufficient depth, without oversimplification?}\\
\texttt{4. **Logic and Coherence**:}\\
   \texttt{- Is the reasoning and logic flow within the section clear and well-structured?}\\
   \texttt{- Are the paragraphs within the section smoothly connected?}\\
\texttt{5. **Language and Expression**:}\\
   \texttt{- Is formal academic language used consistently, avoiding casual expressions?}\\
   \texttt{- Is the language accurate, professional, and aligned with the target domain style?}\\
   \texttt{[Score]}\\
\texttt{For scoring, refer to the following levels:}\\
\texttt{- **0**: The generated paper’s section is highly inadequate, exhibiting significant errors, omissions, or irrelevant content. It falls far short of academic standards and fails to meet even the basic requirements, rendering it unfit for practical use.}\\
\texttt{- **1**: The generated paper’s section has significant issues that need substantial improvement. It is mostly incomplete or lacks depth and clarity, with noticeable errors, omissions, or rough explanations that hinder understanding and fail to meet academic standards.}\\
\texttt{- **2**: The generated paper’s section is generally complete and functional but has noticeable shortcomings. While it covers most of the necessary content, it may lack sufficient detail, depth, or clarity. Minor gaps in structure, logic, or formatting reduce its academic quality, requiring improvement.}\\
\texttt{- **3**: The generated paper’s section is of high academic quality, with only minor issues such as small gaps in detail, slight formatting inconsistencies, or minor errors. These do not hinder clarity or the overall quality. It effectively meets the requirements with room for minimal refinement.}\\
\texttt{- **4**: The generated paper’s section is fully comprehensive and meets the highest academic standards. It includes all required elements—such as detailed explanations, tables, figures, algorithms, or formulas—and is exceptionally well-structured, clear, and error-free. No improvements are necessary.}\\
\texttt{--------------------------------------------------------------------------------}\\
\texttt{Below is the relevant data:}\\
\texttt{[Input Text]}: \\
\texttt{\{input\_text\}} \\
\texttt{**[Model-Generated Response]**:}\\
\texttt{\{response\_text\}} \\
\texttt{**[Human-Assigned Rating]**:}\\
\texttt{\{human\_score\}} \\
\texttt{**[Output Format]**:}\\
\texttt{```json}\\
\texttt{\{\{ }\\
  \texttt{"Structure\_and\_Format": "Your detailed explanation for the score of this dimension.",}\\
  \texttt{"Content\_Coverage": "Your detailed explanation for the score of this dimension.",}\\
  \texttt{"Detail\_and\_Depth": "Your detailed explanation for the score of this dimension.",}\\
  \texttt{"Logic\_and\_Coherence": "Your detailed explanation for the score of this dimension.",}\\
  \texttt{"Language\_and\_Expression": "Your detailed explanation for the score of this dimension.",}\\
\texttt{\}\}}\\
\texttt{```}\\
 \end{tcolorbox}
 \caption{The prompt of generating local explanation.}
 \label{fig:generate_local_explanatin}
\end{figure*}


\begin{figure*}[h]
 \begin{tcolorbox}[colback=blue!2,colframe=blue!50!black]
 \tiny
 \texttt{You are a helpful AI assistant.}\\
 \texttt{Your task is to analyze and explain the scores assigned to a paper across five evaluation dimensions. }\\
 \texttt{For each dimension, the human evaluator considers both the local scores of all relevant sections and the overall quality of the paper, including factors like global logic and structure, in order to assign a paper-level score. Your task is to analyze the global score in relation to the corresponding local scores for each dimension, providing an explanation of how the final paper-level score is determined.}\\
 \texttt{I will provide you with a paper that is divided into \{group\_num\} groups. For each group, you will receive:}\\
 \texttt{1. The content of the group.}\\
 \texttt{2. A human-assigned score that indicates the quality of the corresponding group. The score is divided into five dimensions, with each dimension’s score ranging from [0, 4].}\\
 \texttt{Each group has 5 dimension scores. These 5 scores represent different evaluation dimensions.}
 \texttt{[Evaluation Dimensions]}\\
 \texttt{1. **Structure and Format**:}\\
   \texttt{- Does the generated paper’s section follow standard academic structure?}\\
   \texttt{- Are numbering, subsection titles, and hierarchy within the section clear?}\\
   \texttt{- Is LaTeX formatting (e.g., tables, formulas, citations) correct?}\\
\texttt{2. **Content Coverage**:}\\
   \texttt{- Does the section comprehensively cover all the essential content specified by the input materials?}\\
\texttt{3. **Detail and Depth**:}\\
   \texttt{- Does the section provide enough detailed information to explain its purpose and content fully?}\\
   \texttt{- Are all relevant aspects of the topic explored in sufficient depth, without oversimplification?}\\
\texttt{4. **Logic and Coherence**:}\\
   \texttt{- Is the reasoning and logic flow within the section clear and well-structured?}\\
   \texttt{- Are the paragraphs within the section smoothly connected?}\\
\texttt{5. **Language and Expression**:}\\
   \texttt{- Is formal academic language used consistently, avoiding casual expressions?}\\
   \texttt{- Is the language accurate, professional, and aligned with the target domain style?}\\
   \texttt{[Score]}\\
\texttt{Scores range from 0 to 4, with the following criteria:}\\
\texttt{- **0**: The generated paper’s section is highly inadequate, exhibiting significant errors, omissions, or irrelevant content. It falls far short of academic standards and fails to meet even the basic requirements, rendering it unfit for practical use.}\\
\texttt{- **1**: The generated paper’s section has significant issues that need substantial improvement. It is mostly incomplete or lacks depth and clarity, with noticeable errors, omissions, or rough explanations that hinder understanding and fail to meet academic standards.}\\
\texttt{- **2**: The generated paper’s section is generally complete and functional but has noticeable shortcomings. While it covers most of the necessary content, it may lack sufficient detail, depth, or clarity. Minor gaps in structure, logic, or formatting reduce its academic quality, requiring improvement.}\\
\texttt{- **3**: The generated paper’s section is of high academic quality, with only minor issues such as small gaps in detail, slight formatting inconsistencies, or minor errors. These do not hinder clarity or the overall quality. It effectively meets the requirements with room for minimal refinement.}\\
\texttt{- **4**: The generated paper’s section is fully comprehensive and meets the highest academic standards. It includes all required elements—such as detailed explanations, tables, figures, algorithms, or formulas—and is exceptionally well-structured, clear, and error-free. No improvements are necessary.}\\
\texttt{}\\
\texttt{Input:}\\
\texttt{<PAPER>}: \\
\texttt{<GROUP>} \\
\texttt{**Group Content:** \{content\}} \\
\texttt{**Group Score:** \{score\}} \\
\texttt{</GROUP>} \\
\texttt{...}\\
\texttt{**Overall Score:** \{global\_score\}} \\
\texttt{</PAPER>}\\
\texttt{Output:}\\
\texttt{```json} \\
\texttt{\{\{}\\
\texttt{"Structure\_and\_Format": "Your detailed explanation for the score of this dimension.",} \\
\texttt{"Content\_Coverage": "Your detailed explanation for the score of this dimension.",}\\
\texttt{"Detail\_and\_Depth": "Your detailed explanation for the score of this dimension.",}\\
\texttt{"Logic\_and\_Coherence": "Your detailed explanation for the score of this dimension.",}\\
\texttt{"Language\_and\_Expression": "Your detailed explanation for the score of this dimension.",}\\
\texttt{\}\}}\\
\texttt{```}\\
\texttt{Key Considerations:}\\
\texttt{* Your explanations should align with the scores provided for each dimension and demonstrate a thorough understanding of the paper’s quality in those areas.}\\
\texttt{* Refer to specific aspects of the content, structure, and formatting when justifying the scores.}\\
\texttt{* Be concise but comprehensive in your reasoning.}\\
\texttt{\}\}}\\
\texttt{```}\\
 \end{tcolorbox}
 \caption{The prompt of generating global explanation.}
 \label{fig:prompt_for_agg_exp}
\end{figure*}


\begin{figure*}[h]
 \begin{tcolorbox}[colback=blue!2,colframe=blue!50!black]
 \tiny
 \texttt{You are an expert at evaluating the quality of paper.}\\
 \texttt{**TASK:**}\\
 \texttt{You are tasked with evaluating sections of the paper generated by language models based on provided materials, such as the core idea of the approach and experimental results. For each example, you will review the specific section and assign an objective rating in terms of quality, accuracy, coherence, and overall effectiveness.}\\
 \texttt{Since reviewing an entire paper can be challenging, both the model-generated papers and the human-written reference paper have been divided into sections. You will only need to evaluate and provide a score for the section currently provided.}\\
 \texttt{**Judge Guidelines:**}\\
 \texttt{When assessing the quality of the model-generated section, please consider the following dimensions:}\\
 \texttt{[Evaluation Dimensions]}
 \texttt{1. **Structure and Format**:}\\
   \texttt{- Does the generated paper’s section follow standard academic structure?}\\
   \texttt{- Are numbering, subsection titles, and hierarchy within the section clear?}\\
   \texttt{- Is LaTeX formatting (e.g., tables, formulas, citations) correct?}\\
\texttt{2. **Content Coverage**:}\\
   \texttt{- Does the section comprehensively cover all the essential content specified by the input materials?}\\
\texttt{3. **Detail and Depth**:}\\
   \texttt{- Does the section provide enough detailed information to explain its purpose and content fully?}\\
   \texttt{- Are all relevant aspects of the topic explored in sufficient depth, without oversimplification?}\\
\texttt{4. **Logic and Coherence**:}\\
   \texttt{- Is the reasoning and logic flow within the section clear and well-structured?}\\
   \texttt{- Are the paragraphs within the section smoothly connected?}\\
\texttt{5. **Language and Expression**:}\\
   \texttt{- Is formal academic language used consistently, avoiding casual expressions?}\\
   \texttt{- Is the language accurate, professional, and aligned with the target domain style?}\\
   \texttt{[Score]}\\
\texttt{When estimating the score, please follow the provided scoring levels:}\\
\texttt{- **0**: The generated paper’s section is highly inadequate, exhibiting significant errors, omissions, or irrelevant content. It falls far short of academic standards and fails to meet even the basic requirements, rendering it unfit for practical use.}\\
\texttt{- **1**: The generated paper’s section has significant issues that need substantial improvement. It is mostly incomplete or lacks depth and clarity, with noticeable errors, omissions, or rough explanations that hinder understanding and fail to meet academic standards.}\\
\texttt{- **2**: The generated paper’s section is generally complete and functional but has noticeable shortcomings. While it covers most of the necessary content, it may lack sufficient detail, depth, or clarity. Minor gaps in structure, logic, or formatting reduce its academic quality, requiring improvement.}\\
\texttt{- **3**: The generated paper’s section is of high academic quality, with only minor issues such as small gaps in detail, slight formatting inconsistencies, or minor errors. These do not hinder clarity or the overall quality. It effectively meets the requirements with room for minimal refinement.}\\
\texttt{- **4**: The generated paper’s section is fully comprehensive and meets the highest academic standards. It includes all required elements—such as detailed explanations, tables, figures, algorithms, or formulas—and is exceptionally well-structured, clear, and error-free. No improvements are necessary.}\\
\texttt{}\\
\texttt{[Human-Evaluation Example]}\\
\texttt{We also provide human-evaluation examples, which are intended solely as reference points for score estimation. Note that the content of these examples is unrelated to the current section being evaluated. The examples serve to illustrate the quality expected for each score, helping you better align with human-assigned scores.}: \\
\texttt{Each human-evaluation example includes the following:}\\
\texttt{* The exemplary model-generated response;} \\
\texttt{* The corresponding human-assigned score.}\\
\texttt{For clarity, each example is enclosed within <EXAMPLE> and </EXAMPLE> tags. } \\
\texttt{Here is the human-evaluation example-1: {examples[0]}}\\
\texttt{Here is the human-evaluation example-2: {examples[1]}}\\
\texttt{Here is the human-evaluation example-3: {examples[2]}}\\
\texttt{Here is the human-evaluation example-4: {examples[3]}}\\
\texttt{Here is the human-evaluation example-5: {examples[4]}}\\
\texttt{Please only output the reason and the corresponding score. The score should falls within the range of [0, 4], and the score should be an integer.}\\
\texttt{**Output Format (JSON):**}\\
\texttt{\{\{}\\
\texttt{  "Structure\_and\_Format": \{\{}\\
\texttt{    "score": <score>,}\\
\texttt{    "reason": "<reason>"}\\
\texttt{  \}\},}\\
\texttt{  "Content\_Coverage": \{\{}\\
\texttt{    "score": <score>,}\\
\texttt{    "reason": "<reason>"}\\
\texttt{  \}\},}\\
\texttt{  "Detail\_and\_Depth": \{\{}\\
\texttt{    "score": <score>,}\\
\texttt{    "reason": "<reason>"}\\
\texttt{  \}\},}\\
\texttt{  "Logical\_and\_Coherence": \{\{}\\
\texttt{    "score": <score>,}\\
\texttt{    "reason": "<reason>"}\\
\texttt{  \}\},}\\
\texttt{  "Language\_and\_Expression": \{\{}\\
\texttt{    "score": <score>,}\\
\texttt{    "reason": "<reason>"}\\
\texttt{  \}\},}\\
\texttt{\}\}}\\
\texttt{**Input Materials Corresponding to the Model-Generated Paper Section**}\\
\texttt{\{input\}}\\
\texttt{**Model-Generated Paper Section to be Judged**}\\
\texttt{\{response\}}
 \end{tcolorbox}
 \caption{Prompt for local judge.}
 \label{fig:prompt_for_local_judge}
\end{figure*}


\begin{figure*}[h]
 \begin{tcolorbox}[colback=blue!2,colframe=blue!50!black]
 \tiny
 \texttt{You are an expert at evaluating the quality of academic papers.}\\
 \texttt{Task: Evaluate the Overall Score for Each Dimension of the Paper. }\\
 \texttt{Note that the paper has been divided into several groups, and local scores have already been assigned to each group. When estimating the overall score of the paper, you should take into account both the local scores for each group and the global quality features.}\\
 \texttt{First, I will provide you with {N} human evaluation examples. Each example contains:}\\
 \texttt{1. **Group-Level Scores**: Human ratings for each group across five evaluation dimensions (on a scale of 0-4). Review these scores in detail.}\\
 \texttt{2. **Explanations of Group-Level Scores**: Explanations justifying the scores for each group. Study these explanations thoroughly.}\\
 \texttt{3. **Paper-Level Scores**: Overall ratings for the entire paper across the five evaluation dimensions (on a scale of 0-4), reflecting its comprehensive quality. }
 \texttt{4. **Explanations of Paper-Level Scores**: Explanations justifying the overall scores for each dimension of the paper.}\\
 \texttt{[Evaluation Dimensions]}
 \texttt{1. **Structure and Format**:}\\
   \texttt{- Does the generated paper’s section follow standard academic structure?}\\
   \texttt{- Are numbering, subsection titles, and hierarchy within the section clear?}\\
   \texttt{- Is LaTeX formatting (e.g., tables, formulas, citations) correct?}\\
\texttt{2. **Content Coverage**:}\\
   \texttt{- Does the section comprehensively cover all the essential content specified by the input materials?}\\
\texttt{3. **Detail and Depth**:}\\
   \texttt{- Does the section provide enough detailed information to explain its purpose and content fully?}\\
   \texttt{- Are all relevant aspects of the topic explored in sufficient depth, without oversimplification?}\\
\texttt{4. **Logic and Coherence**:}\\
   \texttt{- Is the reasoning and logic flow within the section clear and well-structured?}\\
   \texttt{- Are the paragraphs within the section smoothly connected?}\\
\texttt{5. **Language and Expression**:}\\
   \texttt{- Is formal academic language used consistently, avoiding casual expressions?}\\
   \texttt{- Is the language accurate, professional, and aligned with the target domain style?}\\
   \texttt{[Score]}\\
\texttt{When estimating the score, please follow the provided scoring levels:}\\
\texttt{- **0**: Highly inadequate; fails to meet basic academic standards.}\\
\texttt{- **1**: Significant issues; incomplete, unclear, or error-prone.}\\
\texttt{- **2**: Adequate but with noticeable shortcomings in depth, detail, or clarity.}\\
\texttt{- **3**: High quality with minor issues; meets requirements effectively.}\\
\texttt{- **4**: Exceptional; comprehensive, clear, and meets the highest academic standards.}\\
\texttt{}\\
\texttt{Examples:}\\
\texttt{\{demonstrations\}}: \\
\texttt{Here is the paper you need to evaluate. It is also divided into \{group\_num\} groups. }\\
\texttt{Each group contains:} \\
\texttt{1. **Group-Level Content**: The content of the paper in the group.}\\
\texttt{2. **Group-Level Scores**: Model ratings for each group across five evaluation dimensions (on a scale of 0-4). Review these scores in detail.} \\
\texttt{3. **Explanations of Group-Level Scores**: Explanations justifying the scores for each group. Study these explanations thoroughly.}\\
\texttt{Please assess the overall score for each dimension of the paper based on its group-level scores and explanations, as well as the paper-level scores and explanations.
\{paper\}}\\
\texttt{Output:}\\
\texttt{Your output must be in JSON format and should include scores for the five dimensions of the paper, along with the corresponding reason for each dimension. Each dimension’s score must be a single value that reflects the paper’s overall performance in that dimension. The score for each dimension should be a number between 0 and 4. Please determine the overall score of the paper in each dimension based on the previous human examples. Use the structure below:}\\
\texttt{```json}\\
\texttt{\{\{}\\
\texttt{  "Structure\_and\_Format": \{\{}\\
\texttt{    "score": <score>,}\\
\texttt{    "reason": "<reason>"}\\
\texttt{  \}\},}\\
\texttt{  "Content\_Coverage": \{\{}\\
\texttt{    "score": <score>,}\\
\texttt{    "reason": "<reason>"}\\
\texttt{  \}\},}\\
\texttt{  "Detail\_and\_Depth": \{\{}\\
\texttt{    "score": <score>,}\\
\texttt{    "reason": "<reason>"}\\
\texttt{  \}\},}\\
\texttt{  "Logical\_and\_Coherence": \{\{}\\
\texttt{    "score": <score>,}\\
\texttt{    "reason": "<reason>"}\\
\texttt{  \}\},}\\
\texttt{  "Language\_and\_Expression": \{\{}\\
\texttt{    "score": <score>,}\\
\texttt{    "reason": "<reason>"}\\
\texttt{  \}\},}\\
\texttt{\}\}}\\
\texttt{```}
 \end{tcolorbox}
 \caption{Prompt for global judge.}
 \label{fig:prompt_for_aggregation}
\end{figure*}

\end{document}